\documentclass[10pt,twocolumn,letterpaper]{article}

\usepackage{cvpr}
\usepackage{times}
\usepackage{epsfig}
\usepackage{graphicx}
\usepackage{amsmath}
\usepackage{amssymb}
\usepackage{algpseudocode}
\usepackage{cases}
\usepackage{subfigure}
\usepackage{overpic}
\usepackage{multirow}
\usepackage{color}
\usepackage{array}
\usepackage{tabularx}
\usepackage[linesnumbered,ruled]{algorithm2e}
\usepackage{bbding}

\usepackage{color}

\usepackage{booktabs}
\usepackage{xcolor}
\usepackage{hhline}
\usepackage{makecell}
\usepackage{colortbl}
\usepackage[normalem]{ulem}
\usepackage{soul}
\usepackage{url}
\definecolor{Gray}{rgb}{0.7,0.7,0.7}

\usepackage[pagebackref=true,breaklinks=true,letterpaper=true,colorlinks,bookmarks=false]{hyperref}

\newcolumntype{x}{>\small c}
\newcolumntype{L}[1]{>{\raggedright\let\newline\\\arraybackslash\hspace{0pt}}m{#1}}
\newcolumntype{C}[1]{>{\centering\let\newline\\\arraybackslash\hspace{0pt}}m{#1}}
\newcolumntype{R}[1]{>{\raggedleft\let\newline\\\arraybackslash\hspace{0pt}}m{#1}}

\newcolumntype{o}{>\small L}

\cvprfinalcopy 

\def\cvprPaperID{***} 

\ifcvprfinal\fi
\begin{document}

\title{Random Erasing Data Augmentation}

\author{Zhun Zhong$^{\dag \S}$, Liang Zheng$^{\S}$, Guoliang Kang$^{\S}$, Shaozi Li$^{\dag}$, Yi Yang$^{\S}$ \\
 {$^{\dag}$Cognitive Science Department, Xiamen University, China} \\
 {$^{\S}$University of Technology Sydney}\\ 
{\tt\small \{zhunzhong007,liangzheng06,yee.i.yang\}@gmail.com} \\
{\tt\small Guoliang.Kang@student.uts.edu.au~  szlig@xmu.edu.cn} 
}

\maketitle

\begin{abstract}
   In this paper, we introduce Random Erasing, a new data augmentation method for training the convolutional neural network (CNN). 
In training, Random Erasing randomly selects a rectangle region in an image and erases its pixels with random values. In this process, training images with various levels of occlusion are generated, which reduces the risk of over-fitting and makes the model robust to occlusion. Random Erasing is parameter learning free, easy to implement, and can be integrated with most of the CNN-based recognition models. Albeit simple, Random Erasing is complementary to commonly used data augmentation techniques such as random cropping and flipping, and yields consistent improvement over strong baselines in image classification, object detection and person re-identification. Code is available at: \url{https://github.com/zhunzhong07/Random-Erasing}.

   
\end{abstract}

\section{Introduction}
      

The ability to generalize is a research focus for the convolutional neural network (CNN). When a model is excessively complex, such as having too many parameters compared to the number of training samples, over-fitting might happen and weaken its generalization ability. 
A learned model may describe random error or noise instead of the underlying data distribution \cite{zhang2016understanding}. In bad cases, the CNN model may exhibit good performance on the training data, but fail drastically when predicting new data. To improve the \emph{generalization ability} of CNNs, many data augmentation and regularization approaches have been proposed, such as random cropping \cite{alexnet}, flipping \cite{vgg}, dropout \cite{srivastava2014dropout}, and batch normalization \cite{ioffe2015batch}.   

Occlusion is a critical influencing factor on the \emph{generalization ability} of CNNs. It is desirable that invariance to various levels of occlusion is achieved. When some parts of an object are occluded, a strong classification model should be able to recognize its category from the overall object structure. However, the collected training samples usually exhibit limited variance in occlusion. 
In an extreme case when all the training objects are clearly visible, \emph{i.e.,} no occlusion happens, the learned CNN will probably work well on testing images without occlusion, but, due to the limited generalization ability of the CNN model, may fail to recognize objects which are partially occluded.
    While we can manually add occluded natural images to the training set, it is costly and the levels of occlusion might be limited. 
    
\begin{figure}[t]
\centering
\includegraphics[width=0.9\linewidth]{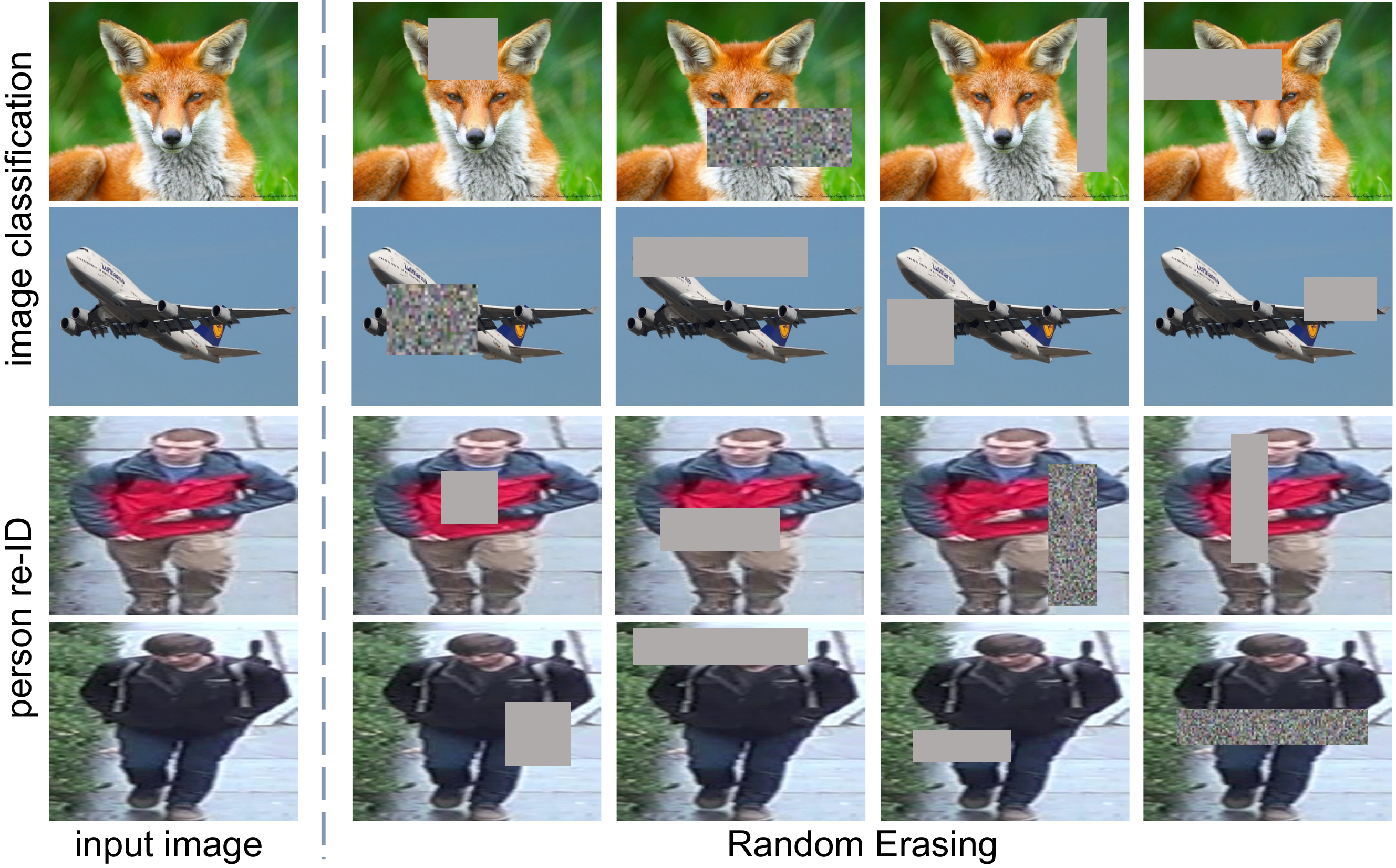}
\caption{Examples of Random Erasing. In CNN training, we randomly choose a rectangle region in the image and erase its pixels with random values or the ImageNet mean pixel value. Images with various levels of occlusion are thus generated.}
\label{fig:example}
\end{figure}

To address the occlusion problem and improve the generalization ability of CNNs, this paper introduces a new data augmentation approach, Random Erasing. It can be easily implemented in most existing CNN models. In the training phase, an image within a mini-batch randomly undergoes either of the two operations: 1) kept unchanged; 2) we randomly choose a rectangle region of an arbitrary size, and assign the pixels within the selected region with random values (or the ImageNet \cite{deng2009imagenet} mean pixel value)\footnote{In Section \ref{Classification Evaluation}, we show erasing with random values achieves approximately equal performance to the ImageNet mean pixel value.}. During Operation 2), an image is partially occluded in a random position with a random-sized mask. 
In this manner, augmented images with various occlusion levels can be generated. Examples of Random Erasing are shown in Fig.~\ref{fig:example}. 
    
Two commonly used data augmentation approaches, \emph{i.e.,}  random flipping and random cropping, also work on the image level and are closely related to Random Erasing. Both techniques have demonstrated the ability to improve the image recognition accuracy. In comparison with Random Erasing, random flipping does not incur information loss during augmentation. Different from random cropping, in Random Erasing, 1) only part of the object is occluded and the overall object structure is preserved, 2)  pixels of the erased region are re-assigned with random values, which can be viewed as adding block noise to the image. 

Working primarily on the fully connected (FC) layer, Dropout \cite{srivastava2014dropout} is also related to our method. It prevents over-fitting by discarding (both hidden and visible) units of the CNN with a probability $p$. Random Erasing is somewhat similar to performing Dropout on the image level. The difference is that in Random Erasing, 1) we operate on a continuous rectangular region,  2) no pixels (units) are discarded, and 3) we focus on making the model more robust to noise and occlusion. The recent A-Fast-RCNN \cite{A-fast-rcnn} proposes an occlusion invariant object detector by training an adversarial network that generates examples with occlusion. Comparison with A-Fast-RCNN, Random Erasing does not require any parameter learning, can be easily applied to other CNN-based recognition tasks and still yields competitive accuracy with A-Fast-RCNN in object detection.
To summarize, Random Erasing has the following advantages:
\begin{itemize}
\item A lightweight method that does not require any extra parameter learning or memory consumption. It can be integrated with various CNN models without changing the learning strategy.
\item A complementary method to existing data augmentation and regularization approaches. When combined, Random Erasing further improves the recognition performance.
\item Consistently improving the performance of recent state-of-the-art deep models on image classification, object detection, and person re-ID. 
\item Improving the robustness of CNNs to partially occluded samples. When we randomly adding occlusion to the CIFAR-10 testing dataset, Random Erasing significantly outperforms the baseline model.
\end{itemize}

\section{Related Work}


Regularization is a key component in preventing over-fitting in the training of CNN models. Various regularization methods have been proposed \cite{alexnet,dropconnect,ba2013adaptive,zeiler2013stochastic,xie2016disturblabel, kang2017patchshuffle}. Dropout \cite{alexnet} randomly discards (setting to zero) the output of each hidden neuron with a probability during the training and only considers the contribution of the remaining weights in forward pass and back-propagation. 
Latter, Wan \emph{et al.} \cite{dropconnect} propose a generalization of dropout approach, DropConect, which instead randomly selects weights to zero during training. In addition, Adaptive dropout  \cite{ba2013adaptive} is proposed where the dropout probability for each hidden neuron is estimated through a binary belief network. Stochastic Pooling \cite{zeiler2013stochastic} randomly selects activation from a multinomial distribution during training, which is parameter free and can be applied with other regularization techniques. Recently, a regularization method named ``DisturbLabel'' \cite{xie2016disturblabel} is introduced by adding noise at the loss layer. DisturbLabel randomly changes the labels of small part of samples to incorrect values during each training iteration. PatchShuffle \cite{kang2017patchshuffle} randomly shuffles the pixels within each local patch while maintaining nearly the same global structures with the original ones, it yields rich local variations for training of CNN.

Data augmentation is an explicit form of regularization that is also widely used in the training of deep CNN \cite{alexnet,vgg, resnet}.
It aims at artificially enlarging the training dataset from existing data using various translations, such as, translation, rotation, flipping, cropping, adding noises, \emph{etc}. 
The two most popular and effective data augmentation methods in training of deep CNN are
random flipping and random cropping. Random flipping randomly flips the input image horizontally, while random cropping extracts random sub-patch from the input image.
As an analogous choice, Random Erasing may discard some parts of the object. For random cropping, it may crop off the corners of the object, while Random Erasing may occlude some parts of the object. Random Erasing maintains the global structure of object. Moreover, it can be viewed as adding noise to the image. The combination of random cropping and Random Erasing can produce more various training data. Recently, Wang \emph{et al.} \cite{A-fast-rcnn} learn an adversary with Fast-RCNN \cite{fast-rcnn} detection to create hard examples on the fly by blocking some feature maps spatially. Instead of generating occlusion examples in feature space, Random Erasing generates images from the original images with very little computation which is in effect, computationally free and does not require any extra parameters learning.

\section{Datasets}
For \textbf{image classification}, we evaluate on three image classification datasets, including two well-known datasets, \textbf{CIFAR-10} and \textbf{CIFAR-100} \cite{cifar}, and a new dataset  \textbf{Fashion-MNIST} \cite{xiao2017fashion}. \textbf{CIFAR-10} and \textbf{CIFAR-100} contain 50,000 training and 10,000 testing 32$\times$32 color images drawn from 10 and 100 classes, respectively. \textbf{Fashion-MNIST} consists of 60,000 training and 10,000 testing 28x28 gray-scale images. Each image is associated with a label from 10 classes. We evaluate top-1 error rates in the format ``mean $\pm$ std'' based on 5 runs.

For \textbf{object detection}, we use the \textbf{PASCAL VOC 2007} \cite{everingham2010pascal} dataset which contains 9,963 images of 24,640 annotated objects in training/validation and testing sets. We use the ``trainval'' set for training and ``test'' set for testing.  

For \textbf{person re-identification (re-ID)}, the \textbf{Market-1501} dataset \cite{zheng2015scalable} contains 32,668 labeled bounding boxes of 1,501 identities captured from 6 different cameras.  The dataset is split into two parts: 12,936 images with 751 identities for training and 19,732 images with 750 identities for testing. In testing, 3,368 hand-drawn images with 750 identities are used as probe set to identify the correct identities on the testing set. \textbf{DukeMTMC-reID} \cite{zheng2017unlabeled,ristani2016MTMC} contains 36,411 images of 1,812 identities shot by 8 high-resolution cameras. Similar to Market-1501, it contains 16,522 training images of 702 identities, 2,228 query images of the other 702 identities and 17,661 gallery images. \textbf{CUHK03} \cite{CUHK03} contains 14,096 images of 1,467 identities. We use the \textbf{new training/testing protocol} proposed in  \cite{zhong2017re} to evaluate the multi-shot re-ID performance. There are 767 identities in the training set and 700 identities in the testing set. We conduct experiment on both ``detected'' and ``labeled'' sets. We evaluate using rank-1 accuracy and mean average precision (mAP) on these three datasets.

\section{Our Approach}
       This section presents the Random Erasing data augmentation method for training the convolutional neural network (CNN). We first describe the detailed procedure of Random Erasing. Next, the implementation of Random Erasing in different tasks is introduced. Finally, we analyze the differences between Random Erasing and random cropping.
       
\subsection{Random Erasing}
In training, Random Erasing is conducted with a certain probability. For an image $I$ in a mini-batch, the probability of it undergoing Random Erasing is $p$, and the probability of it being kept unchanged is $1 - p$.
In this process, training images with various levels of occlusion are generated. 
    
    Random Erasing randomly selects a rectangle region $I_e$ in an image, and erases its pixels with random values. Assume that the size of the training image is $W \times H$. The area of the image is $S=W \times H$. We randomly initialize the area of erasing rectangle region to $S_e$, where $\frac{S_e}{S}$ is in range specified by minimum $s_l$ and maximum $s_h$. The aspect ratio of erasing rectangle region is randomly initialized between $r_1$ and $r_2$, we set it to $r_e$. The size of $I_e$ is $H_e = \sqrt{S_e \times r_e}$ and $W_e = \sqrt{\frac{S_e}{r_e}}$.
    Then, we randomly initialize a point $\mathcal{P} = (x_e, y_e)$ in $I$. If $x_e + W_e \le W$ and $y_e + H_e \le H$, we set the region, $I_e = (x_e, y_e, x_e+W_e, y_e+H_e)$, as the selected rectangle region. Otherwise repeat the above process until an appropriate $I_e$ is selected. With the selected erasing region $I_e$, each pixel in $I_e$ is assigned to a random value in [0, 255], respectively. The procedure of selecting the rectangle area and erasing this area is shown in Alg. \ref{algorithm 1}.


\begin{algorithm}[t]
\SetAlgoLined
\SetKwInOut{Input}{Input}
\SetKwInOut{Output}{Output}
\SetKwInput{Initialization}{Initialization}
\caption{Random Erasing Procedure}\label{algorithm 1}
\Input{Input image $I$; \\
       Image size $W$ and $H$; \\
       Area of image $S$; \\
       Erasing probability $p$; \\
       Erasing area ratio range $s_l$ and $s_h$; \\      
       Erasing aspect ratio range $r_1$ and $r_2$.}
\Output{Erased image $I^{\ast}$.}
\Initialization{$p_1 \leftarrow $ Rand (0, 1).} 

\eIf{$p_1 \geq p$}{
   $I^{\ast} \leftarrow I$; \\
   \Return{$I^{\ast}$}.
}{
   \While {True}{
      $S_e\leftarrow $ Rand $(s_l, s_h)$$\times S$;\\
      $r_e \leftarrow $ Rand $(r_1, r_2)$;\\
      $H_e \leftarrow \sqrt{S_e \times r_e}$,~ $W_e \leftarrow \sqrt{\frac{S_e}{r_e}}$;\\
      $x_e \leftarrow $ Rand $(0, W)$,~ $y_e \leftarrow $ Rand $(0, H)$;\\
      \If{$x_e + W_e \le W$ and $y_e + H_e \le H$}{
          $I_e \leftarrow (x_e, y_e, x_e+W_e, y_e+H_e)$;\\
          $I(I_e) \leftarrow $ Rand (0, 255);\\
          $I^{\ast} \leftarrow I$;\\
          \Return{$I^{\ast}$}.
          }
     }
}
\end{algorithm}

\subsection{Random Erasing for Image Classification and Person Re-identification}

In image classification, an image is classified according to its visual content. In general, training data does not provide the location of the object, so we could not know where the object is. In this case,   we perform Random Erasing on the whole image according to Alg. \ref{algorithm 1}.

Recently, the person re-ID model is usually trained in a classification network for embedding learning \cite{reid-survey}. In this task, since pedestrians are confined with detected bounding boxes, persons are roughly in the same position and take up the most area of the image. In this scenario, we adopt the same strategy as image classification, as in practice, the pedestrian can be occluded in any position. We randomly select rectangle regions on the whole pedestrian image and erase it. Examples of Random Erasing for image classification and person re-ID are shown in Fig. \ref{fig:example}.

\subsection{Random Erasing for Object Detection}
\begin{figure}[!t]
\centering
\includegraphics[width=0.9\linewidth]{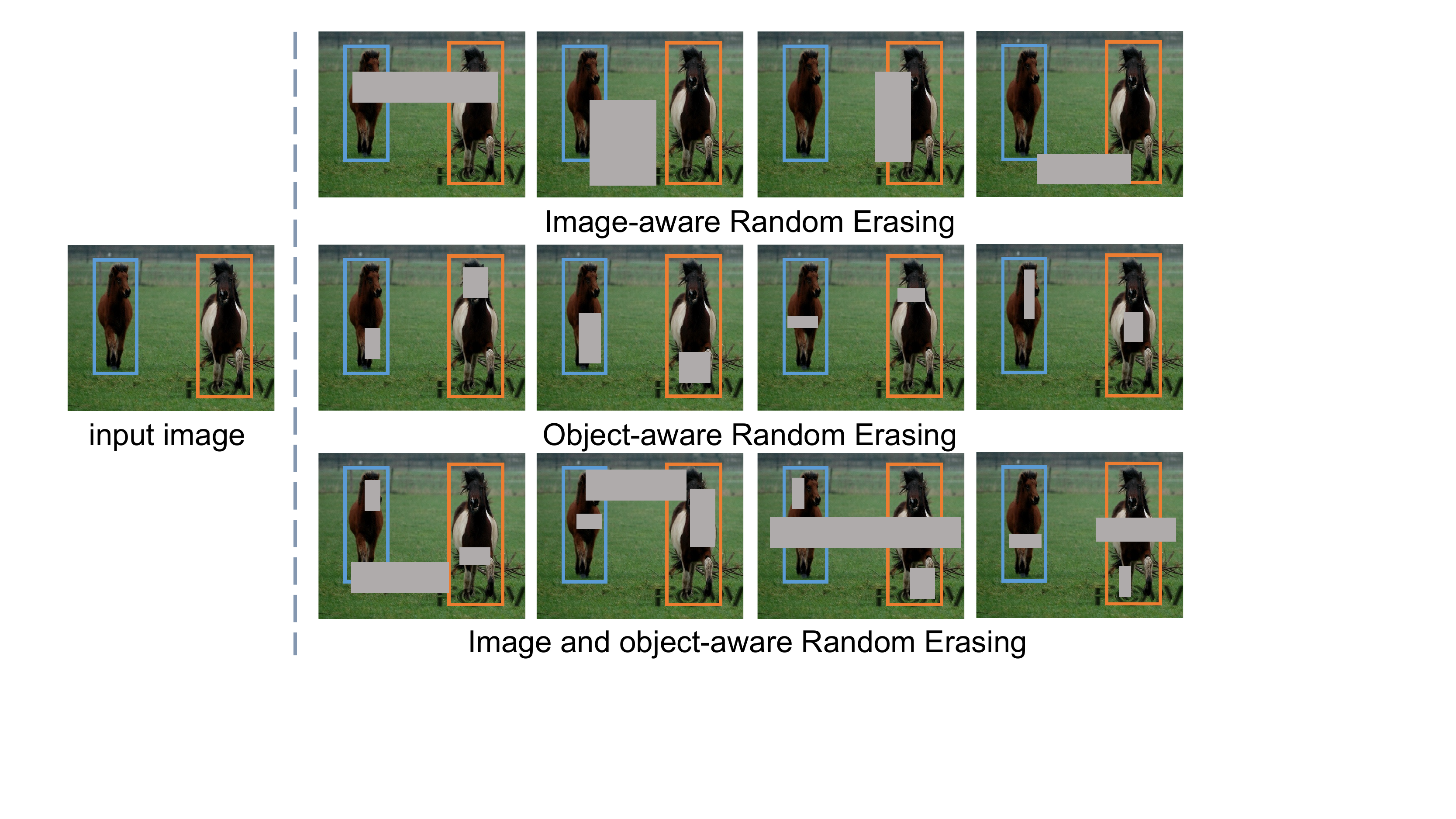}
\caption{Examples of Random Erasing for object detection with Image-aware Random Erasing (\textbf{IRE}), Object-aware Random Erasing (\textbf{ORE}) and Image and object-aware Random Erasing (\textbf{I+ORE}).}
\label{fig:example_detection}
\end{figure}

Object detection aims at detecting instances of semantic objects of a certain class in images. Since the location of each object in the training image is known, we implement Random Erasing with three schemes:1) Image-aware Random Erasing (\textbf{IRE}): selecting erasing regions on the whole image, the same as image classification and person re-identification; 2) Object-aware Random Erasing (\textbf{ORE}): selecting erasing regions in the bounding box of each object. In the latter, if there are multiple objects in the image, Random Erasing is applied on each object separately. 3) Image and object-aware Random Erasing (\textbf{I+ORE}): selecting erasing regions in both the whole image and each object bounding box. Examples of Random Erasing for object detection with the three schemes are shown in Fig. \ref{fig:example_detection}. 

\subsection{Comparison with Random Cropping}
\begin{figure}[!t]
\centering
\includegraphics[width=0.9\linewidth]{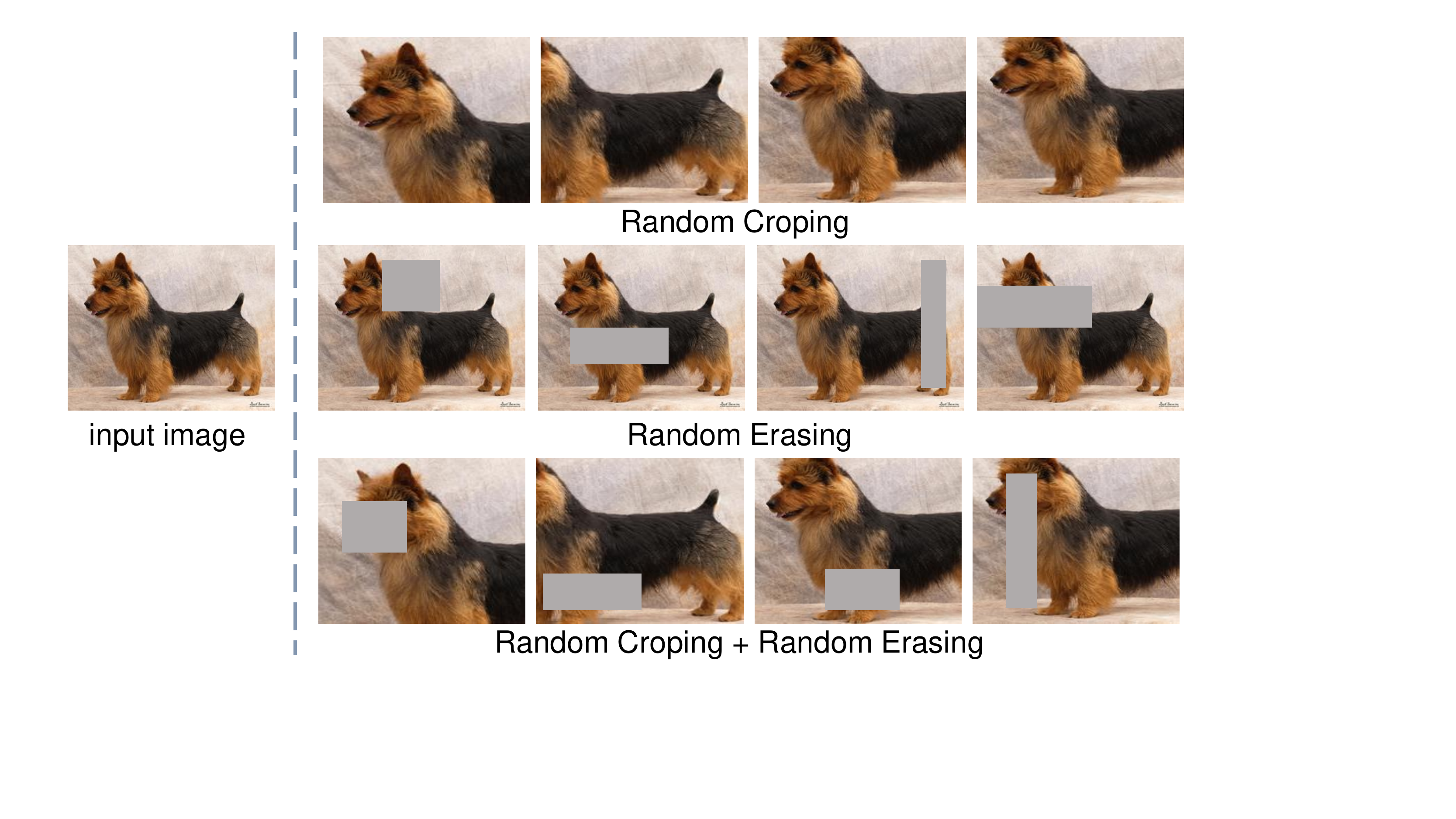}
\caption{Examples of Random Erasing, random cropping, and the combination of them. When combining these two augmentation methods, more various images can be generated.}
\label{fig:example_randomcrop}
\end{figure}

    Random cropping is an effective data augmentation approach, it reduces the contribution of the background in the CNN decision, and can base learning models on the presence of parts of the object instead of focusing on the whole object. In comparison to random cropping, Random Erasing retains the overall structure of the object, only occluding some parts of object. In addition, the pixels of erased region are re-assigned with random values, which can be viewed as adding noise to the image.
    In our experiment (Section \ref{Classification Evaluation}), we show that these two methods are complementary to each other for data augmentation.
    The examples of Random Erasing, random cropping, and the combination of them are shown in Fig. \ref{fig:example_randomcrop}.

\section{Experiment}

\begin{table*}
\footnotesize
\begin{center}
\newcolumntype{C}{>{\centering\arraybackslash}X}%
\newcolumntype{R}{>{\raggedleft\arraybackslash}X}%
\newcolumntype{L}{>{\raggedright\arraybackslash}X}%
\begin{tabularx}{\linewidth}{ l|CC|CC|CC }
\hline
\multirow{2}{*}{Model}  & \multicolumn{2}{c|}{CIFAR-10}  & \multicolumn{2}{c|}{CIFAR-100}  & \multicolumn{2}{c}{Fashion-MNIST} \\
\cline{2-7} 
 & Baseline & Random Erasing & Baseline & Random Erasing & Baseline & Random Erasing\\
\hline
\hline
ResNet-20 & 7.21 $\pm$ 0.17&  6.73 $\pm$ 0.09 & 30.84 $\pm$ 0.19 & 29.97 $\pm$ 0.11 & 4.39 $\pm$ 0.08 & 4.02 $\pm$ 0.07 \\
ResNet-32 & 6.41 $\pm$ 0.06 &  5.66 $\pm$ 0.10 &  28.50 $\pm$ 0.37 &  27.18 $\pm$ 0.32 &  4.16 $\pm$ 0.13 & 3.80 $\pm$ 0.05 \\
ResNet-44 & 5.53 $\pm$ 0.08 &  5.13 $\pm$ 0.09 &  25.27 $\pm$ 0.21 &  24.29 $\pm$ 0.16 & 4.41 $\pm$ 0.09 &  4.01 $\pm$ 0.14 \\
ResNet-56 & 5.31 $\pm$ 0.07 &  4.89 $\pm$ 0.07 &  24.82 $\pm$ 0.27 & 23.69 $\pm$ 0.33 &  4.39 $\pm$ 0.10 &  4.13 $\pm$ 0.42 \\
ResNet-110 & 5.10 $\pm$ 0.07 &  4.61 $\pm$ 0.06 &  23.73 $\pm$ 0.37 & 22.10 $\pm$ 0.41 &  4.40 $\pm$ 0.10 & 4.01 $\pm$ 0.13 \\  
\hline
\hline
ResNet-20-PreAct & 7.36 $\pm$ 0.11 & 6.78 $\pm$ 0.06 & 30.58 $\pm$ 0.16 & 30.18 $\pm$ 0.13 & 4.43 $\pm$ 0.19
 &  4.02 $\pm$ 0.09 \\
ResNet-32-PreAct & 6.42 $\pm$ 0.11 &  5.79 $\pm$ 0.10 & 29.04 $\pm$ 0.25 & 27.82 $\pm$ 0.28 & 4.36 $\pm$ 0.02 & 4.00 $\pm$ 0.05 \\
ResNet-44-PreAct & 5.54 $\pm$ 0.16 &  5.09 $\pm$ 0.10 &  25.22 $\pm$ 0.19 & 24.10 $\pm$ 0.26 &  4.92 $\pm$ 0.30 &  4.23 $\pm$ 0.15 \\
ResNet-56-PreAct & 5.28 $\pm$ 0.12 &  4.84 $\pm$ 0.09 & 24.14 $\pm$ 0.25 & 22.93 $\pm$ 0.27 &  4.55 $\pm$ 0.30 &  3.99 $\pm$ 0.08 \\
ResNet-110-PreAct & 4.80 $\pm$ 0.09 & 4.47 $\pm$ 0.11 &  22.11 $\pm$ 0.20 & 20.99 $\pm$ 0.11 &  5.11 $\pm$ 0.55 & 4.19 $\pm$ 0.15 \\
\hline
\hline
ResNet-18-PreAct & 5.17	$\pm$ 0.18 & 4.31 $\pm$ 0.07 & 24.50 $\pm$ 0.29 & 24.03 $\pm$ 0.19 &  4.31 $\pm$ 0.06 &  3.90 $\pm$ 0.06 \\
WRN-28-10 & 3.80 $\pm$ 0.07 &  \textbf{3.08  $\pm$ 0.05} & \textbf{18.49 $\pm$ 0.11} &  \textbf{17.73 $\pm$ 0.15} & \textbf{4.01 $\pm$ 0.10} &  \textbf{3.65 $\pm$ 0.03} \\
ResNeXt-8-64 & \textbf{3.54 $\pm$ 0.04} & 3.24 $\pm$ 0.03 &  19.27 $\pm$ 0.30 & 18.84 $\pm$ 0.18 & 4.02 $\pm$ 0.05 & 3.79 $\pm$ 0.06 \\
\hline
\end{tabularx}
\end{center}
\vspace{-.1in}
\caption{\label{tabel:classification result}Test errors (\%) with different architectures on CIFAR-10, CIFAR-100 and Fashion-MNIST.}
\end{table*}

\begin{figure*}[!t]
\centering
\subfigure[probability $p$]{\includegraphics[width=0.33\linewidth]{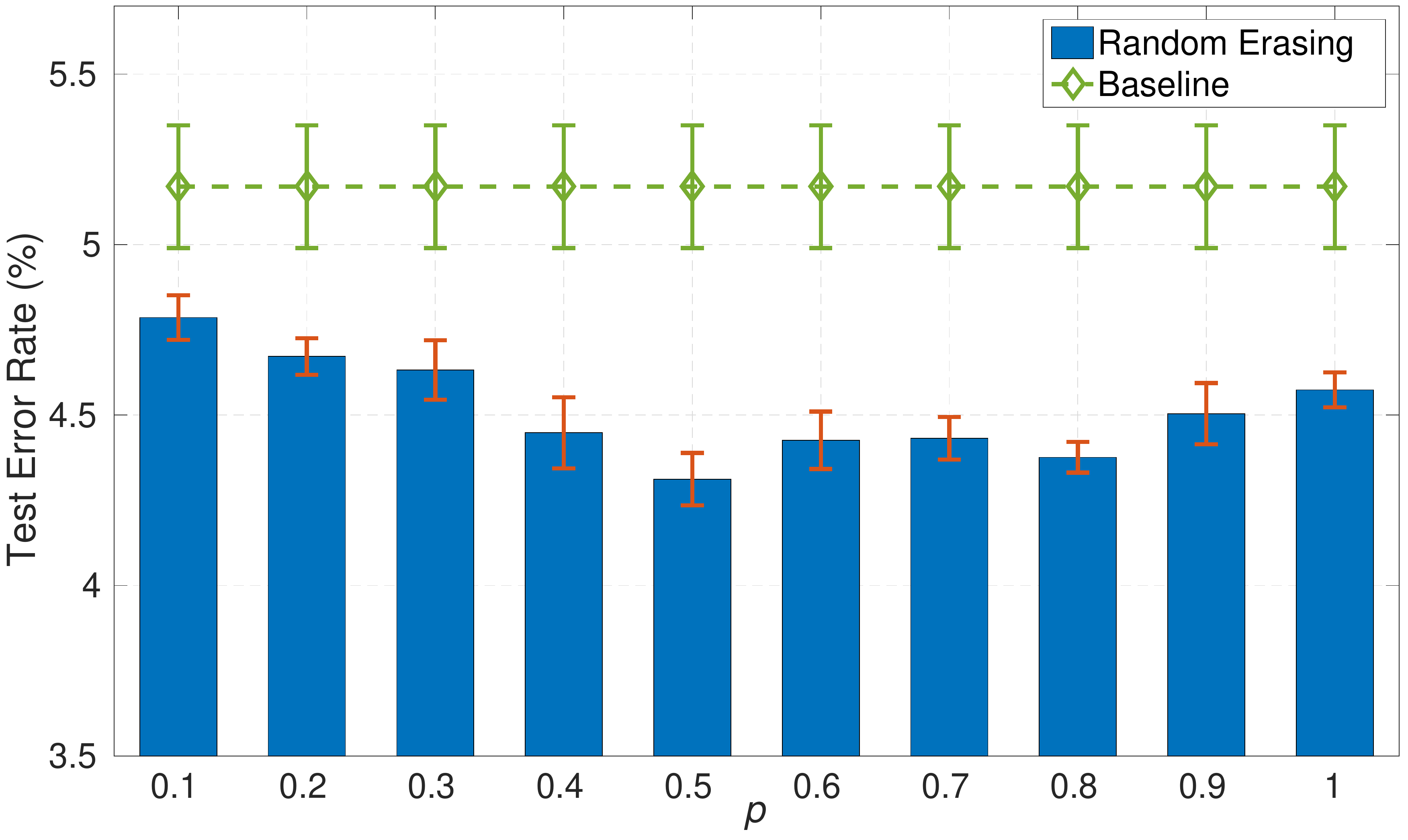}}
\subfigure[area ratio $s_h$]{\includegraphics[width=0.33\linewidth]{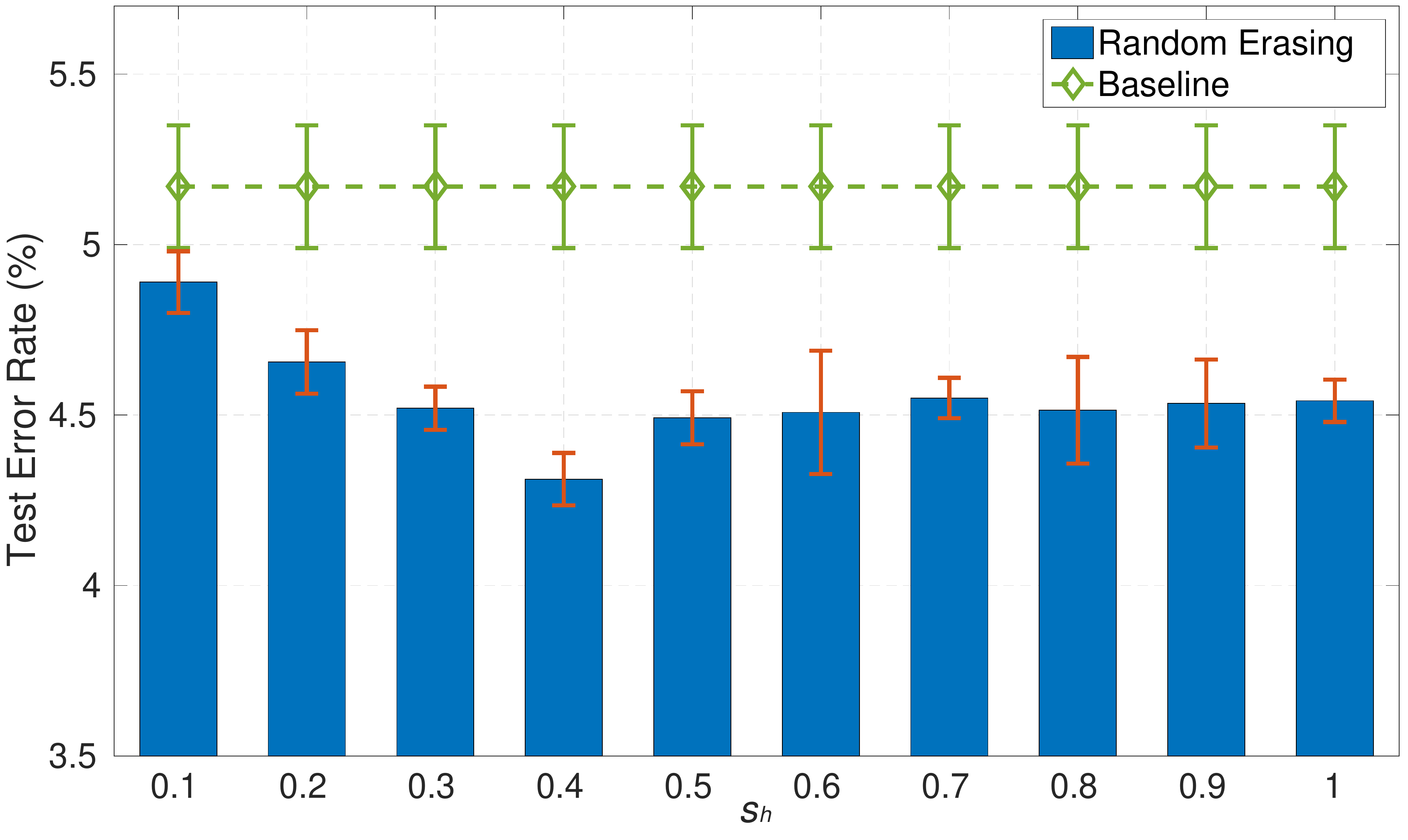}}
\subfigure[aspect ratio $r_1$]{\includegraphics[width=0.33\linewidth]{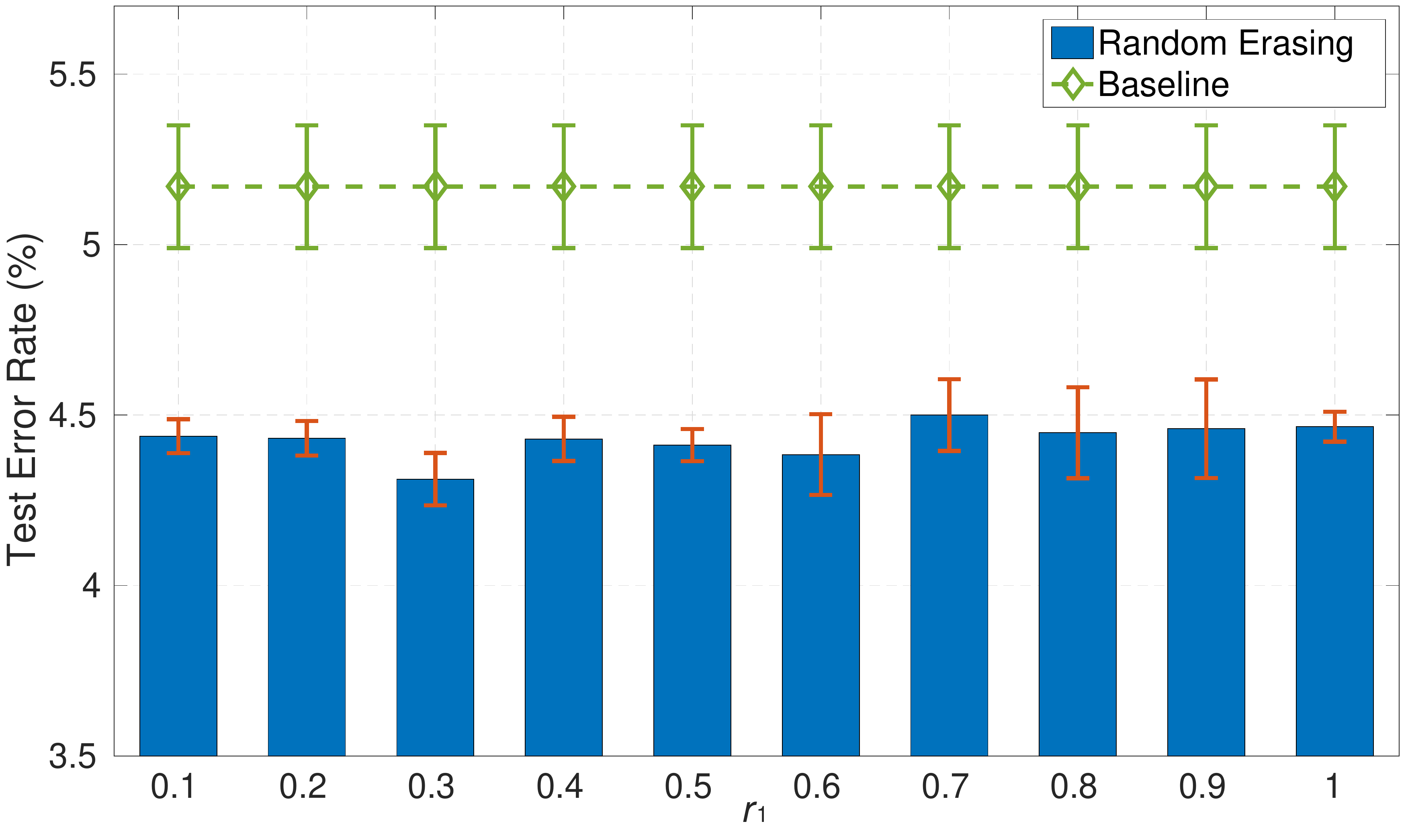}}
\vspace{-.02in}
\caption{Test errors (\%) under different hyper-parameters on CIFAR-10 with using ResNet18 (pre-act).}
\label{fig:CIFAR-parameter}
\end{figure*}

\begin{table*}
\footnotesize
\begin{center}
\newcolumntype{C}{>{\centering\arraybackslash}X}%
\newcolumntype{R}{>{\raggedleft\arraybackslash}X}%
\begin{tabularx}{\linewidth}{ l|C|C|C|C|C}
\hline
Types of Erasing Value & Baseline & RE-R & RE-M & RE-0 & RE-255  \\
\hline
\hline
Test error rate (\%)& 5.17 $\pm$ 0.18 & \textbf{4.31 $\pm$ 0.07} & 4.35 $\pm$ 0.12 & 4.62 $\pm$ 0.09 & 4.85 $\pm$ 0.13\\
\hline
\end{tabularx}
\end{center}
\vspace{-.1in}
\caption{\label{tabel:Erasing Value} Test errors (\%) on CIFAR-10 based on ResNet18 (pre-act) with four types of erasing value. \textbf{Baseline:} Baseline model, \textbf{RE-R:} Random Erasing model with random value, \textbf{RE-M:} Random Erasing model with mean value of ImageNet 2012, \textbf{RE-0:} Random Erasing model with 0, \textbf{RE-255:} Random Erasing model with 255.}
\end{table*}

\subsection{Image Classification}

\subsubsection{Experiment Settings} In all of our experiment, we compare the CNN models trained with or without Random Erasing. For the same deep architecture, all the models are trained from the same weight initialization. Note that some popular regularization techniques (\eg, weight decay, batch normalization and dropout) and various data augmentations (\eg, flipping, padding and cropping) are employed. The compared CNN architectures are summarized as below:

\textbf{Architectures.} Four architectures are adopted on CIFAR-10, CIFAR-100 and Fashion-MNIST: ResNet \cite{resnet}, pre-activation ResNet \cite{he2016identity}, ResNeXt \cite{xie2016aggregated}, and Wide Residual Networks \cite{zagoruyko2016wide}. 
We use the 20, 32, 44, 56, 110-layer network for ResNet and pre-activation ResNet. The 18-layer network is also adopted for pre-activation ResNet. We use ResNeXt-29-8$\times$64 and WRN-28-10 in the same way as \cite{xie2016aggregated} and \cite{zagoruyko2016wide}, respectively.  
The training procedure follows \cite{resnet}. Specially, the learning rate
starts from 0.1 and is divided by 10 after the 150th and 225th epoch. We stop training by the 300th epoch. If not specified, all models are trained with data augmentation: randomly performs horizontal flips, and takes a random crop with 32$\times$32 for CIFAR-10 and CIFAR-100 (28$\times$28 for Fashion-MNIST) from images padded by 4 pixels on each side.

\subsubsection{Classification Evaluation}
\label{Classification Evaluation}
\textbf{Classification accuracy on different datasets.}
The results of applying Random Erasing on CIFAR-10 ,CIFAR-100 and Fashion-MNIST with different architectures are shown in Table~\ref{tabel:classification result}. We set $p = 0.5$, $s_l = 0.02$, $s_h = 0.4$, and $r_1 = \frac{1}{r_2} = 0.3$. Results indicate that models trained with Random Erasing have significant improvement, demonstrating that our method is applicable to various CNN architectures. For CIFAR-10, our method improves the accuracy by 0.49\% and 0.33\% using ResNet-110 and ResNet-110-PreAct, respectively. In particular, our approach obtains 3.08\% error rate  using WRN-28-10, which improves the accuracy by 0.72\% and achieves new state of the art. For CIFAR-100, our method obtains 17.73\% error rate which gains 0.76\% than the WRN-28-10 baseline. Our method  also works well for gray-scale images: Random erasing improves WRN-28-10 from 4.01\% to 3.65\% in top-1 error on Fashion-MNIST.

\textbf{The impact of hyper-parameters.} When implementing Random Erasing on CNN training, we have three hyper-parameters to evaluate, \emph{i.e.}, the erasing probability $p$, the area ratio range of erasing region $s_l$  and $s_h$, and the aspect ratio range of erasing region $r_1$ and $r_2$. To demonstrate the impact of these hyper-parameters on the model performance, we conduct experiment on CIFAR-10 based on ResNet18 (pre-act) under varying hyper-parameter settings. To simplify experiment, we fix $s_l$ to 0.02, $r_1 = \frac{1}{r_2}$ and evaluate $p$, $s_h$, and $r_1$. We set $p=0.5$, $s_h=0.4$ and $r_1=0.3$ as the base setting.  When evaluating one of the parameters, we fixed the other two parameters. Results are shown in Fig. \ref{fig:CIFAR-parameter}.

Notably, Random Erasing consistently outperforms the ResNet18 (pre-act) baseline under all parameter settings. For example, when $p \in [0.2, 0.8]$ and $s_h \in [0.2, 0.8]$, the average classification error rate is $4.48 \%$, outperforming the baseline method ($5.17 \%$) by a large margin. Random Erasing is also robust to the aspect ratios of the erasing region. Specifically, our best result (when $r_1=0.3$, error rate = $4.31 \%$) reduces the classification error rate by 0.86\% compared with the baseline. 
In the following experiment for image classification, we set $p = 0.5$, $s_l = 0.02$, $s_h = 0.4$, and $r_1 = \frac{1}{r_2} = 0.3$, if not specified.


\textbf{Four types of random values for erasing.} 
We evaluate Random Erasing when pixels in the selected region are erased in four ways: 1) each pixel is assigned with a random value ranging in [0, 255], denoted as RE-R; 2) all pixels are assign with the mean ImageNet pixel value \emph{i.e.,} [125, 122, 114], denoted as RE-M; 3) all pixels are assigned with 0, denoted as RE-0; 4) all pixels are assigned with 255, denoted as RE-255. Table~\ref{tabel:Erasing Value} presents the result with different erasing values on CIFAR10 using ResNet18 (pre-act). We observe that, 1) all erasing schemes outperform the baseline, 2) RE-R achieves approximately equal performance to RE-M, and 3) both RE-R and RE-M are superior to RE-0 and RE-255. If not specified, we use RE-R in the following experiment.


\begin{table}
\footnotesize
\begin{center}
\newcolumntype{C}{>{\centering\arraybackslash}X}%
\newcolumntype{R}{>{\raggedleft\arraybackslash}X}%
\newcolumntype{L}{>{\raggedright\arraybackslash}X}%
\begin{tabularx}{\linewidth}{ >{\scriptsize}l|>{\scriptsize}C|| >{\scriptsize}l|>{\scriptsize}C }
\hline
Method & Test error (\%) & Method & Test error (\%) \\
\hline
\hline
Baseline & 5.17 $\pm$  0.18 & Baseline & 5.17 $\pm$  0.18\\
Random Erasing & \textbf{4.31 $\pm$ 0.07} & Random Erasing & \textbf{4.31 $\pm$ 0.07} \\
\hline 
\hline
Dropout & Test error (\%) & Random Noise & Test error (\%) \\
\hline 
\hline 
$\lambda_1 = 0.001$ & 5.37 $\pm$ 0.12 & $\lambda_2 = 0.01$ & 5.38 $\pm$ 0.07 \\ 
$\lambda_1 = 0.005$ & 5.48 $\pm$ 0.15 & $\lambda_2 = 0.05$ & 5.79 $\pm$ 0.14\\ 
$\lambda_1 = 0.01$ & 5.89 $\pm$ 0.14 & $\lambda_2 = 0.1$ & 6.13 $\pm$ 0.12\\ 
$\lambda_1 = 0.05$ & 6.23 $\pm$ 0.11 & $\lambda_2 = 0.2$ & 6.25 $\pm$ 0.09\\ 
$\lambda_1 = 0.1$ & 6.38 $\pm$ 0.18 & $\lambda_2 = 0.4$ & 6.52 $\pm$ 0.12\\
\hline 
\end{tabularx}
\end{center}
\vspace{-.1in}
\caption{\label{tabel:dropout and noise} Comparing Random Erasing with dropout and random noise on CIFAR-10 with using ResNet18 (pre-act).}
\end{table}


\textbf{Comparison with Dropout and random noise.} 
We compare Random Erasing with two variant methods applied on image layer. 1) Dropout: we apply dropout on image layer with probability $\lambda_1$. 2) Random noise: we add different levels of noise on the input image by changing the pixel to a random value in [0, 255] with probability $\lambda_2$. The probability of whether an image undergoes dropout or random noise is set to 0.5 as Random Erasing. Results are presented in Table \ref{tabel:dropout and noise}. 
It is clear that applying dropout or adding random noise at the image layer fails to improve the accuracy. As the probability $\lambda_1$ and $\lambda_2$ increase, performance drops quickly. When $\lambda_2 = 0.4$, the number of noise pixels for random noise is approximately equal to the number of erasing pixels for Random Erasing, the error rate of random noise increases from 5.17\% to 6.52\%, while Random Erasing reduces the error rate to 4.31\%.

\textbf{Comparing with data augmentation methods. }
We compare our method with random flipping and random cropping in Table~\ref{tabel:CIFAR-10-data_augmentation_study}. When applied alone, random cropping (6.33\%) outperforms the other two methods.
Importantly, \textbf{Random Erasing and the two competing techniques are complementary}. Particularly, combining these three methods achieves 4.31\% error rate,  a 7\% improvement over the baseline without any augmentation. 

\begin{figure}[!t]
\centering
\includegraphics[width=\linewidth]{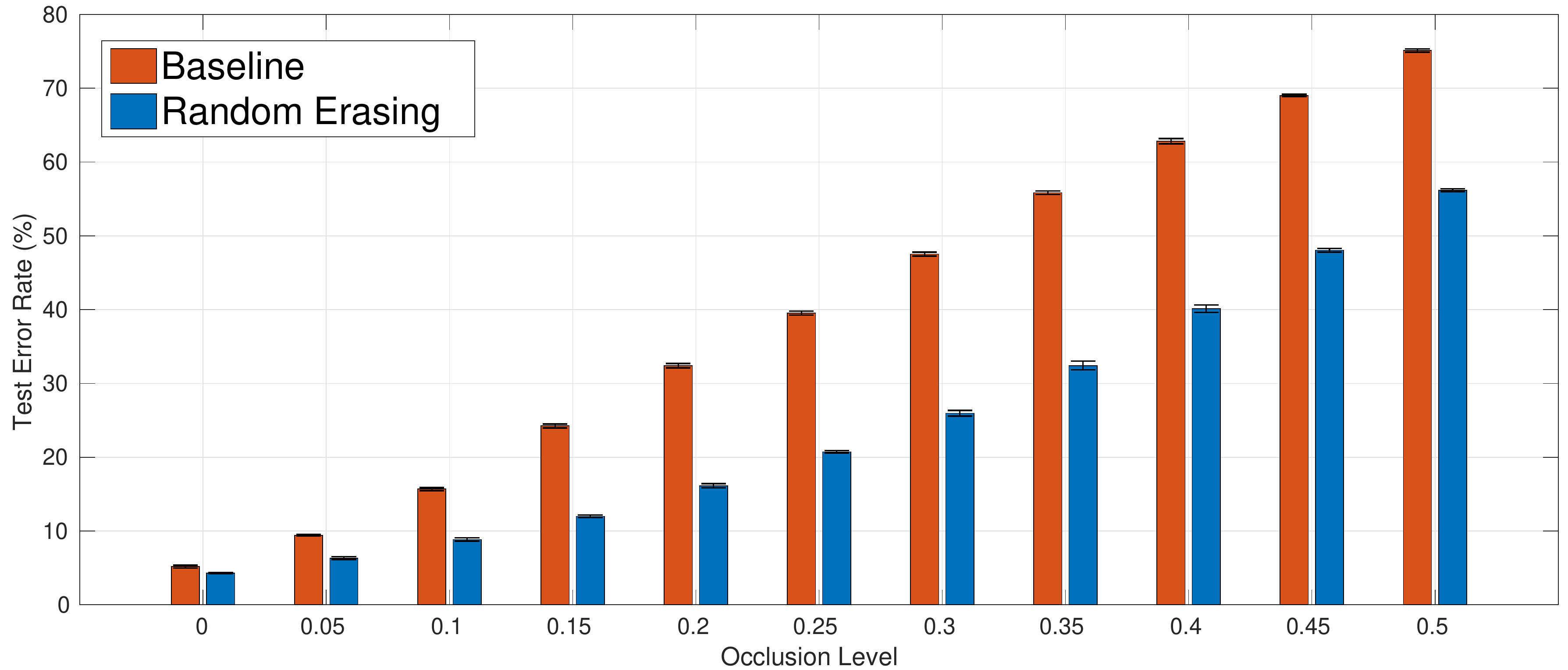}
\caption{Test errors (\%) under different levels of occlusion on CIFAR-10 based on ResNet18 (pre-act).}
\label{fig:CIFAR-10-occlusion}
\end{figure}

\textbf{Robustness to occlusion}.
Last, we show the robustness of Random Erasing against occlusion. In this experiment, we add different levels of occlusion to the CIFAR-10 dataset in testing. We randomly select a region of area and fill it with random values. The aspect ratio of the region is randomly chosen from the range of [0.3, 3.33]. Results as shown in Fig.~\ref{fig:CIFAR-10-occlusion}. Obviously, the baseline performance drops quickly when increasing the occlusion level $l$. In comparison, the performance of the model training with Random Erasing decreases slowly. Our approach achieves 56.36\% error rate when the occluded area is half of the image ($l = 0.5$), while the baseline rapidly drops to 75.04\%. It demonstrates that Random Erasing improves the robustness of CNNs against occlusion.

\subsection{Object Detection}
\subsubsection{Experiment Settings} Experiment is conducted based on the Fast-RCNN~\cite{fast-rcnn} detector. The model is initialized by the ImageNet classification models, and then fine-tuned on the object detection data. We experiment with VGG16 \cite{vgg} architecture. We follow A-Fast-RCNN \cite{A-fast-rcnn} for training. We apply SGD for 80K to train all models. The training rate starts with 0.001 and decreases to 0.0001 after 60K iterations. With this training procedure, the baseline mAP is slightly better than the report mAP in \cite{fast-rcnn}. We use the selective search proposals during training. For  Random Erasing, we set $p = 0.5$, $s_l = 0.02$, $s_h = 0.2$, and $r_1 = \frac{1}{r_2} = 0.3$. 


\begin{table}
\footnotesize
\begin{center}
\newcolumntype{C}{>{\centering\arraybackslash}X}%
\newcolumntype{R}{>{\raggedleft\arraybackslash}X}%
\begin{tabularx}{\linewidth}{ l||C|C|C||c }
\hline
Method  & RF & RC & RE & Test errors (\%) \\
\hline
\hline
\multirow{8}{*}{Baseline}  &  &  &  & 11.31 $\pm$ 0.18 \\
& $\checkmark$ &   &  & 8.30 $\pm$ 0.17 \\
&  & $\checkmark$  &  & 6.33 $\pm$ 0.15 \\
&  &  & $\checkmark$ &  10.13 $\pm$ 0.14 \\
& $\checkmark$ & $\checkmark$ &  & 5.17 $\pm$ 0.18\\
& $\checkmark$ &  & $\checkmark$ & 7.19 $\pm$ 0.10 \\
&  & $\checkmark$ & $\checkmark$ & 5.21 $\pm$ 0.14 \\
& $\checkmark$ & $\checkmark$ & $\checkmark$ & \textbf{4.31} $\pm$ \textbf{0.07} \\
\hline
\end{tabularx}
\end{center}
\vspace{-.1in}
\caption{\label{tabel:CIFAR-10-data_augmentation_study} Test errors (\%) with different data augmentation methods on CIFAR-10 based on ResNet18 (pre-act). \textbf{RF}: Random flipping, \textbf{RC}: Random cropping, \textbf{RE}: Random Erasing.}
\end{table}

\begin{table*}[t]
\centering

\renewcommand{\arraystretch}{1.2}
\renewcommand{\tabcolsep}{1.2mm}
\resizebox{\linewidth}{!}{

\begin{tabular}{@{}L{2.1cm} !{\color{gray}\vrule} L{1.2cm} |   l | r*{19}{x} @{}}
\hline
Method  & train set & mAP & aero      & bike      & bird      & boat      & bottle     & bus        & car        & cat        & chair      & cow        & table      & dog        & horse      & mbike      & persn     & plant      & sheep      & sofa       & train      & tv   \\
\hline
\hline
FRCN~\cite{fast-rcnn}  &07 & 66.9 & 74.5 & 78.3 & 69.2 & 53.2 & 36.6 & 77.3 & 78.2 & 82.0 & 40.7 & 72.7 & 67.9 & 79.6 & 79.2 & 73.0 & 69.0 & 30.1 & 65.4 & 70.2 & 75.8 & 65.8 \\
FRCN\raisebox{0.2ex}{$\star$}~\cite{A-fast-rcnn} &07  & 69.1 & 75.4 & 80.8 & 67.3 & 59.9 & 37.6 & 81.9 & 80.0 & 84.5 & 50.0 & 77.1 & 68.2 & 81.0 & 82.5 & 74.3 & 69.9 & 28.4 & 71.1 & 70.2 & 75.8 & 66.6  \\
ASDN~\cite{A-fast-rcnn}  &07  & 71.0 & 74.4 & 81.3 & 67.6 & 57.0 & 46.6 & 81.0 & 79.3 & 86.0 & 52.9 & 75.9 & 73.7 & 82.6 & 83.2 & 77.7 & 72.7 & 37.4 & 66.3 & 71.2 & 78.2 & 74.3 \\
Ours (IRE)  &07  & 70.5  & 75.9  & 78.9  & 69.0   & 57.7  & 46.4  & 81.7  & 79.5  & 82.9  & 49.3  & 76.9  & 67.9  & 81.5  & 83.3  & 76.7  & 73.2  & 40.7  & 72.8  & 66.9  & 75.4  & 74.2   \\
Ours (ORE)  &07  & 71.0 & 75.1 & 79.8 & 69.7 & 60.8 & 46.0 & 80.4 & 79.0 & 83.8 & 51.6 & 76.2 & 67.8 & 81.2 & 83.7 & 76.8 & 73.8 & 43.1 & 70.8 & 67.4 & 78.3 & 75.6  \\
Ours (I+ORE)  &07  & \textbf{71.5}  & 76.1 & 81.6 & 69.5 & 60.1 & 45.6 & 82.2 & 79.2 & 84.5 & 52.5 & 78.7 & 71.6 & 80.4 & 83.3 & 76.7 & 73.9 & 39.4 & 68.9 & 69.8 & 79.2 & 77.4 \\
\hline
\hline
FRCN~\cite{fast-rcnn}  &07+12 &70.0 &  77.0 &
78.1 &69.3 &59.4 &38.3 &81.6 &78.6 &86.7 &42.8 &78.8 &68.9 &84.7 &82.0 &76.6 &69.9 &31.8 &70.1 &74.8 &80.4 &70.4  \\
FRCN\raisebox{0.2ex}{$\star$}~\cite{A-fast-rcnn} &07+12  & 74.8 & 78.5  & 81.0  & 74.7  & 67.9  & 53.4  & 85.6  & 84.4  & 86.2  & 57.4  & 80.1  & 72.2  & 85.2  & 84.2  & 77.6  & 76.1  & 45.3  & 75.7  & 72.3  & 81.8  & 77.3   \\
Ours (IRE)  &07+12  & 75.6 & 79.0 & 84.1 & 76.3 & 66.9 & 52.7 & 84.5 & 84.4 & 88.7 & 58.0 & 82.9 & 71.1 & 84.8 & 84.4 & 78.6 & 76.7 & 45.5 & 77.1 & 76.3 & 82.5 & 76.8 \\
Ours (ORE)  &07+12 & 75.8 & 79.4 & 81.6 & 75.6 & 66.5 & 52.7 & 85.5 & 84.7 & 88.3 & 58.7 & 82.9 & 72.8 & 85.0 & 84.3 & 79.3 & 76.3 & 46.3 & 76.3 & 74.9 & 86.0 & 78.2 \\
Ours (I+ORE)  &07+12 & \textbf{76.2} & 79.6 & 82.5 & 75.7 & 70.5 & 55.1 & 85.2 & 84.4 & 88.4 & 58.6 & 82.6 & 73.9 & 84.2 & 84.7 & 78.8 & 76.3 & 46.7 & 77.9 & 75.9 & 83.3 & 79.3  \\

\hline
\end{tabular}
}
\vspace{-.05in}
\caption[caption]{ {\bf VOC 2007 test} detection average precision (\%).  FRCN\raisebox{0.2ex}{$\star$} refers to FRCN with training schedule in ~\cite{A-fast-rcnn}.}
\label{tab:voc2007}
\end{table*}


\begin{table*}
\footnotesize
\begin{center}
\newcolumntype{C}{>{\centering\arraybackslash}X}%
\newcolumntype{R}{>{\raggedleft\arraybackslash}X}%
\begin{tabularx}{\linewidth}{ l|c|c|CC|CC|CC|CC }
\hline
\multirow{2}{*}{Method}  & \multirow{2}{*}{Model}  & \multirow{2}{*}{RE} &  \multicolumn{2}{c|}{Market-1501} & \multicolumn{2}{c|}{DukeMTMC-reID} & \multicolumn{2}{c|}{CUHK03 (labeled)} & \multicolumn{2}{c}{CUHK03 (detected)} \\
\cline{4-11} 
 & & & Rank-1 & mAP & Rank-1 & mAP & Rank-1 & mAP & Rank-1 & mAP \\
\hline
\hline
 \multirow{6}{*}{IDE}  & \multirow{2}{*}{ResNet-18} & No & 79.87&57.37&67.73& 46.87&28.36&25.65&26.86&25.04 \\
 &   & Yes & 82.36&62.06&70.60& 51.41&36.07&32.58&34.21& 31.20 \\
\cline{2-11}
 &  \multirow{2}{*}{ResNet-34} & No & 82.93&62.34&71.63& 49.71&31.57&28.66&30.14& 27.55 \\
  &   & Yes & 84.80&65.68&73.56& 54.46&40.29& 35.50&36.36& 33.46 \\
  \cline{2-11}
 &  \multirow{2}{*}{ResNet-50} & No & 83.14&63.56&71.99&51.29&30.29& 27.37&28.36& 26.74 \\
  &   & Yes & 85.24&68.28&74.24& 56.17&41.46& 36.77&38.50& 34.75 \\
  \cline{2-11}
\hline
\hline
 \multirow{6}{*}{TriNet}  & \multirow{2}{*}{ResNet-18} & No & 77.32& 58.43&67.50& 46.27&43.00&39.16&40.50& 37.36 \\
 &   & Yes & 79.84& 61.68&71.81& 51.84&48.29&43.80&46.57& 43.20\\
 \cline{2-11}
 &  \multirow{2}{*}{ResNet-34} & No & 80.73& 62.65 &72.04& 51.56&46.00& 43.79&45.07& 42.58 \\
  &   & Yes & 83.11& 65.98&72.89&55.38&53.07& 48.80&53.21& 48.03 \\
  \cline{2-11}
 &  \multirow{2}{*}{ResNet-50} & No & 82.60& 65.79&72.44& 53.50&49.86& 46.74&50.50& 46.47 \\
  &   & Yes & 83.94& 68.67&72.98& 56.60&\textbf{58.14}&\textbf{53.83}&\textbf{55.50}&\textbf{50.74} \\
\hline\hline
 \multirow{2}{*}{SVDNet}  & \multirow{2}{*}{ResNet-50} & No & 84.41& 65.60&76.82& 57.70&42.21&38.73&41.85&38.24 \\
 &   & Yes & \textbf{87.08}& \textbf{71.31}&\textbf{79.31} & \textbf{62.44}&49.43& 45.07&48.71& 43.50\\
 \cline{2-11}
  \cline{2-11}
\hline

\end{tabularx}
\end{center}
\vspace{-.1in}
\caption{\label{tabel:reid} Person re-identification performance with Random Erasing (RE) on Market-1501, DukeMTMC-reID, and CUHK03 based on different models. We evaluate CUHK03 under the new evaluation protocol in \cite{zhong2017re}. }
\end{table*}

\subsubsection{Detection Evaluation}

We report results with using IRE, ORE and I+ORE during training Fast-RCNN in Table~\ref{tab:voc2007}. The detector is trained with two training set, VOC07 trainval and union of VOC07 and VOC12 trainval. When training with VOC07 trainval, the baseline is 69.1\% mAP. The detector learned with IRE scheme achieves an improvement to 70.5\% mAP and the ORE scheme obtains 71.0\% mAP. The ORE performs slightly better than IRE. When implementing Random Erasing on  overall image and objects, the detector training with I+ORE obtains further improved in performance with 71.5\% mAP. Our approach (I+ORE) outperforms A-Fast-RCNN \cite{A-fast-rcnn} by 0.5\% in mAP. Moreover, our method does not require any parameter learning and is easy to implement. When using the enlarged 07+12 training set, the baseline is 74.8\% which is much better than only using 07 training set. The IRE and ORE schemes give similar results, in which the mAP of IRE is improved by 0.8\% and ORE is improved by 1.0\%. When applying I+ORE during training, the mAP of Fast-RCNN increases to 76.2\%, surpassing the baseline by 1.4\%.

\subsection{Person Re-identification}

\subsubsection{Experiment Settings}
Three baselines are used in person re-ID, \ie, the ID-discriminative Embedding (IDE) \cite{reid-survey}, TriNet \cite{hermans2017defense}, and SVDNet \cite{sun2017svdnet}. IDE and SVDNet are trained with the Softmax loss, while TriNet is trained with the triplet loss. The input images are resized to 256 $\times$ 128. 

For IDE, we basically follow the training strategy in \cite{reid-survey}. We further add a fully connected layer with 128 units after the Pool5 layer, followed by batch normalization, ReLU and Dropout. The Dropout probability is set to 0.5. We use SGD to train IDE. The learning rate starts with 0.01 and is divided by 10 after each 40 epochs. We train 100 epochs in total. In testing, we extract the output of Pool5 as feature for Market-1501 and DukeMTMC-reID datasets, and the fully connected layer with 128 units as feature for CUHK03. 

For TriNet and SVDNet, we use the same model as proposed in \cite{hermans2017defense} and \cite{sun2017svdnet}, respectively, and follow the same training strategy. In testing, we extract the last fully connected layer with 128 units as feature for TriNet and extract the output of Pool5 for SVDNet. Note that, we use 256 $\times$ 128 as the input size to train SVDNet which achieves higher performance than the original paper using size 224 $\times$ 224.

We use the ResNet-18, ResNet-34, and ResNet-50 architectures for IDE and TriNet, and ResNet-50 for SVDNet. We fine-tune them on the model pre-trained on ImageNet \cite{deng2009imagenet}. We also perform random cropping and random horizontal flipping during training. For Random Erasing, we set $p = 0.5$, $s_l = 0.02$, $s_h = 0.2$, and $r_1 = \frac{1}{r_2} = 0.3$.

\subsubsection{Person Re-identification Performance}
\textbf{Baseline Evaluation.} The results of Random Erasing on Market-1501, DukeMTMC-reID, and CUHK03 with different baselines and architectures are shown in Table~\ref{tabel:reid}. For Market-1501 and DukeMTMC-reID, the IDE \cite{reid-survey} and SVDNet \cite{sun2017svdnet} baselines outperform the TriNet baseline \cite{hermans2017defense}. Since there exists plenty of samples in each ID, the models with using the Softmax loss can learn better features. Specially, the IDE achieves 83.14\% and 71.99\% rank-1 accuracy on Market-1501 and DukeMTMC-reID with using ResNet-50, respectively. SVDNet gives rank-1 accuracy of 84.41\% and 76.82\% on Market-1501 and DukeMTMC-reID with ResNet-50, respectively. This is 1.81\% higher for Market-1501 and 4.38\% higher for DukeMTMC-reID than the TriNet with ResNet-50.
However, on CUHK03, the performance of TriNet is higher than IDE and SVDNet, since the lack of training samples compromises the  Softmax loss.  TriNet obtains 49.86\% rank-1 accuracy and 46.74\% mAP on CUHK03 for the labeled setting with ResNet-50.


\begin{table}
\small
\begin{center}
\newcolumntype{C}{>{\centering\arraybackslash}X}%
\newcolumntype{R}{>{\raggedleft\arraybackslash}X}%
\begin{tabularx}{\linewidth}{ l|C|C }
\hline
Method & Rank-1 & mAP \\
\hline
\hline
{BOW}  \cite{zheng2015scalable}  & 34.40 & 14.09 \\
LOMO+XQDA \cite{liao2015lomo} & 43.79 & 22.22 \\
DNS \cite{zhang2016learningDNS}  & 61.02 & 35.68 \\
Gated \cite{varior2016gated}  & 65.88 & 39.55 \\
IDE \cite{reid-survey} & 72.54 & 46.00\\
MSCAN \cite{li2017learning} & 80.31 & 57.53\\
DF \cite{zhao2017deeply} & 81.0 & 63.4 \\
SSM \cite{bai2017scalable} & 82.21 & 68.80 \\
{SVDNet} \cite{sun2017svdnet} & 82.3 & 62.1 \\
{GAN} \cite{zheng2017unlabeled} & 83.97 & 66.07 \\
PDF \cite{su2017pose}  & 84.14 & 63.41 \\
TriNet \cite{hermans2017defense} & 84.92 & \textcolor{green}{69.14} \\
DJL \cite{li2017person} & \textcolor{green}{85.1} & 65.5\\
\hline
\hline
SVDNet+Ours&  \textcolor{blue}{87.08}& \textcolor{blue}{71.31} \\
SVDNet+Ours+re \cite{zhong2017re}& \textcolor{red}{89.13}& \textcolor{red}{83.93}\\
\hline
\end{tabularx}
\end{center}
\vspace{-.1in}
\caption{\label{tabel:Market-state} Comparison of our method with state-of-the-art methods on the Market-1501 dataset. We use ResNet-50 as backbone. The best, second and third
highest results are in \textcolor{red}{red}, \textcolor{blue}{blue} and \textcolor{green}{green}, respectively.}
\end{table}

\textbf{Random Erasing improves different baseline models.} When implementing Random Erasing in these baseline models, we can observe that, Random Erasing consistently improves the rank-1 accuracy and mAP. Specifically, for Market-1501, Random Erasing improves the rank-1 by 3.10\% and 2.67\% for IDE and SVDNet with using ResNet-50. For DukeMTMC-reID, Random Erasing increases the rank-1 accuracy from 71.99\% to 74.24\% for IDE (ResNet-50) and from 76.82\% to 79.31\% for SVDNet (ResNet-50). For CUHK03, TriNet gains 8.28\% and 5.0\% in rank-1 accuracy when applying Random Erasing on the labeled and detected settings with ResNet-50, respectively. We note that, due to lack of adequate training data, over-fitting tend to occur on CUHK03. For example,  a deeper architecture, such as ResNet-50,  achieves lower performance than ResNet-34 when using the IDE mode on the detected subset. However, with our method, IDE (ResNet-50) outperforms IDE (ResNet-34). This indicates that our method can reduce the risk of over-fitting and improves the re-ID performance.


\begin{table}
\small
\begin{center}
\newcolumntype{C}{>{\centering\arraybackslash}X}%
\newcolumntype{R}{>{\raggedleft\arraybackslash}X}%
\begin{tabularx}{\linewidth}{ l|C|C }
\hline
Method & Rank-1 & mAP \\
\hline
\hline
{BOW}+kissme  \cite{zheng2015scalable}  & 25.13 & 12.17 \\
LOMO+XQDA \cite{liao2015lomo} & 30.75 & 17.04  \\ 
IDE \cite{reid-survey} & 65.22 & 44.99 \\
{GAN} \cite{zheng2017unlabeled} & 67.68 & 47.13 \\
OIM \cite{xiao2017oim} & 68.1 & 47.4 \\
TriNet \cite{hermans2017defense} & 72.44 & 53.50 \\ 	%
ACRN \cite{schumann2017person} & 72.58 & 51.96\\
SVDNet \cite{sun2017svdnet} & 76.7 & 56.8 \\
\hline 
\hline
SVDNet+Ours & \textbf{79.31} & \textbf{62.44} \\
SVDNet+Ours+re \cite{zhong2017re}& \textbf{84.02}& \textbf{78.28} \\
\hline
\end{tabularx}
\end{center}
\vspace{-.1in}
\caption{\label{tabel:duke-state} Comparison of our method with state-of-the-art methods on the DukeMTMC-reID dataset. We use ResNet-50 as backbone.}
\end{table}


\begin{table}
\small
\begin{center}

\newcolumntype{C}{>{\centering\arraybackslash}X}%
\newcolumntype{R}{>{\raggedleft\arraybackslash}X}%
\begin{tabularx}{\linewidth}{ l|c|C|c|C }
\hline
\multirow{2}{*}{Method} & \multicolumn{2}{c|}{Labeled} & \multicolumn{2}{c}{Detected} \\
\cline{2-5}
  & Rank-1 &  mAP  & Rank-1 & mAP \\
\hline
\hline
BOW+XQDA \cite{zheng2015scalable} & 7.93 & 7.29& 6.36 & 6.39 \\
LOMO+XQDA \cite{liao2015lomo} & 14.8 & 13.6 & 12.8 & 11.5  \\ 
IDE \cite{reid-survey}  & 22.2	& 21.0 & 21.3 & 19.7 \\
IDE+DaF \cite{yu2017divide} & 27.5 &	31.5 &	26.4 & 30.0  \\
SVDNet \cite{sun2017svdnet} & 40.9 & 37.8 & 41.5 & 37.2 \\
DPFL \cite{Chen_2017_ICCV} & 43.0 &	40.5 &	40.7 &	37.0 \\
TriNet \cite{hermans2017defense} & 49.86& 46.74&50.50& 46.47 \\
\hline
\hline
TriNet+Ours&  \textbf{58.14} & \textbf{53.83} &  \textbf{55.50} & \textbf{50.74}\\
TriNet+Ours+re \cite{zhong2017re}&  \textbf{63.93} & \textbf{65.05} &  \textbf{64.43} & \textbf{64.75} \\
\hline
\end{tabularx}
\end{center}
\vspace{-.1in}
\caption{\label{tabel:CUHK03-state} Comparison of our method with state-of-the-art methods on the CUHK03 dataset using the new evaluation protocol in \cite{zhong2017re}. We use ResNet-50 as backbone.}
\end{table}


\textbf{Comparison with the state-of-the-art methods.} We compare our method with the state-of-the-art methods on Market-1501, DukeMTMC-reID, and CUHK03 in Table~\ref{tabel:Market-state}, Table~\ref{tabel:duke-state}, and Table~\ref{tabel:CUHK03-state}, respectively. Our method achieves competitive results with the state of the art. Specifically, 
 based on SVDNet, our method obtains \textbf{rank-1 accuracy = 87.08\%}  for Market-1501, and \textbf{rank-1 accuracy = 79.31\%}  for DukeMTMC-reID. On CUHK03, based on TriNet, our method achieves \textbf{rank-1 accuracy = 58.14\%} for the labeled setting, and \textbf{rank-1 accuracy = 55.50\%} for the detected setting.

When we further combine our system with re-ranking  \cite{zhong2017re,bai2016sparse}, the final rank-1 performance arrives at 89.13\% for Market-1501, 84.02\% for DukeMTMC-reID, and 64.43\% for CUHK03 under the detected setting.



\section{Conclusion}
In this paper, we propose a new data augmentation approach named ``Random Erasing'' for training the convolutional neural network (CNN). It is easy to implemented: Random Erasing randomly occludes an arbitrary region of the input image during each training iteration. Experiment conducted on CIFAR10, CIFAR100, and Fashion-MNIST with various architectures validate the effectiveness of our method. Moreover, we obtain reasonable improvement on object detection and person re-identification, demonstrating that our method has good performance on various recognition tasks. In the future work, we will apply our approach to other CNN recognition tasks, such as, image retrieval and face recognition.

{\small
\bibliographystyle{ieee}
\bibliography{egbib}
}

\end{document}